\documentstyle[aclap,psfig]{article}

\newlength{\captionwidth}
\makeatletter
\long\def\@makecaption#1#2{
\vskip 10pt\setbox\@tempboxa\hbox{\small #1: #2}
\hbox to \hsize{\hfil
\ifdim\wd\@tempboxa>\captionwidth
\hbox to\captionwidth{\hsize=\captionwidth\vbox{\unhbox\@tempboxa\par}}\else
\hbox to\captionwidth{\hfil\box\@tempboxa\hfil}\fi\hfil}}
\makeatother
\def\given{\,|\,}
\def\argmax{\mathop{\rm arg\, max\,}}
\def\argmin{\mathop{\rm arg\, min\,}}

\def\pt{\tilde p}

\def\ww{\sc}

\def\P{P}

\def\V{{\cal W}}
\def\nV{{\mid \V \mid}}
\def\WSJ{{\bf WSJ}}
\def\BN{{\bf BN}}
\def\TDT{{\bf TDT}}
\def\triprob{{p_{\mbox{\scriptsize tri}}}} 
\def\expprob{{p_{\mbox{\scriptsize exp}}}} 
\def\yes{\mbox{\sc yes}}
\def\no{\mbox{\sc no}}
\def\syes{\mbox{\sc yes}}
\def\sno{\mbox{\sc no}}
\def\pt{\tilde p}

\def\given{\,|\,}

\def\W{N}

\title{\LARGE\bf Text Segmentation Using Exponential Models\thanks{Research
supported in part by NSF grant IRI-9314969,
DARPA AASERT award DAAH04-95-1-0475, and the 
ATR Interpreting Telecommunications Research Laboratories.}}
\author{{\bf Doug Beeferman \quad Adam Berger \quad John Lafferty} \\[5pt]
School of Computer Science\\
Carnegie Mellon University}

\begin{document}
\bibliographystyle{fullname}

\maketitle

\begin{abstract}
This paper introduces a new statistical approach to partitioning text
automatically into coherent segments. Our approach enlists both
short-range and long-range language models to help it sniff out likely
sites of topic changes in text. To aid its search, the system consults a
set of simple lexical hints it has learned to associate with the
presence of boundaries through inspection of a large corpus of annotated
data. We also propose a new probabilistically motivated error metric for
use by the natural language processing and information retrieval
communities, intended to supersede precision and recall for appraising
segmentation algorithms.  Qualitative assessment of our algorithm as
well as evaluation using this new metric demonstrate the effectiveness
of our approach in two very different domains, {\it Wall Street
Journal\/} articles and the {\rm TDT Corpus\/}, 
a collection of newswire articles and broadcast news transcripts.
\end{abstract}

\section{Introduction} 
The task we address in this paper might seem on the face of it rather
elementary: identify where one region of text ends and another begins.
This work was motivated by the observations that such a seemingly simple
problem can actually prove quite difficult to automate, and that a tool
for partitioning a stream of undifferentiated text (or multimedia) into
coherent regions would be of great benefit to a number of existing
applications.

The task itself is ill-defined: what exactly is meant by a ``region'' of
text?  We confront this issue by adopting an empirical definition of
segment. At our disposal is a collection of online data ($38$ million
words of Wall Street Journal archives and another $150$ million words
from selected news broadcasts) annotated with the boundaries between
regions---articles or news reports, respectively. Given this input, the
task of constructing a segmenter may be cast as a problem in machine
learning: glean from the data a set of hints about where boundaries
occur, and use these hints to inform a decision on where to place breaks
in unsegmented data.

A general-purpose tool for partitioning expository text or multimedia data into
coherent regions would have a number of immediate practical uses. In fact, this
research was inspired by a problem in information retrieval: given a large
unpartitioned collection of expository text and a user's query, return a
collection of coherent segments matching the query. Lacking a segmenting tool,
an IR application may be able to locate positions in its database which are
strong matches with the user's query, but be unable to determine how much of
the surrounding data to provide to the user. This can manifest itself in quite
unfortunate ways. For example, a video-on-demand application (such as the one
described in \cite{Christel:95}) responding to a query about a recent news
event may provide the user with a news clip related to the event, followed 
or preceded by part of an unrelated story or even a commercial.

Document summarization is another fertile area for an automatic segmenter.
Summarization tools often work by breaking the input into ``topics'' and then
summarizing each topic independently. A segmentation tool has obvious
applications to the first of these tasks. 

The output of a segmenter could also serve as input to various
language-modeling tools. For instance, one could envision segmenting a corpus,
classifying the segments by topic, and then constructing topic-dependent
language models from the generated classes.

The paper will proceed as follows.  In Section~\ref{sec:prevwork} we
very briefly review some previous approaches to the text segmentation
problem.  In Section~\ref{sec:elm} we describe our model, including the
type of linguistic clues it looks for in deciding when placing a
partition is appropriate.  In Section~\ref{sec:fi} we describe a feature
induction algorithm that automatically constructs a set of the most
informative clues.  Section~\ref{sec:construction} shows examples of the
feature induction algorithm in action.  In Section~\ref{sec:error} we
introduce a new, probabilistically motivated way to evaluate a text
segmenter.  Finally, in Section~\ref{sec:results} we demonstrate our
model's effectiveness on two distinct domains.

\section{Some Previous Work}
\label{sec:prevwork}
In this section we very briefly discuss some previous approaches 
to the text segmentation problem.
\subsection{Text tiling}

The {\it TextTiling\/} algorithm, introduced by Hearst \cite{Hearst:94},
segments expository texts into multiple paragraphs of coherent 
discourse units.  
A cosine measure is used to gauge the similarity between constant-size
blocks of morphologically analyzed tokens.  First-order rates of change
of this measure are then calculated to decide the placement of
boundaries between blocks, which are then adjusted to coincide with the
paragraph segmentation, provided as input to the algorithm.
This approach leverages the observation that text segments are dense
with repeated content words.  Relying on this fact, however, may limit
precision because the repetition of concepts within a document is more subtle
than can be recognized by only a ``bag of words'' tokenizer and
morphological filter.  

Word pairs other than ``self-triggers,'' for
example, can be discovered automatically from training data using the
techniques of mutual information employed by our language model.  
Furthermore, Hearst's approach segments at the paragraph level, which may
be too coarse for applications like information retrieval on transcribed
or automatically recognized spoken documents, in which paragraph
boundaries are not known.

\subsection{Lexical cohesion}

\cite{Kozima:93} employs a ``lexical cohesion profile'' to keep
track of the semantic cohesiveness of words in a text within a fixed-length
window.  In contrast to Hearst's focus on strict repetition, Kozima uses
a semantic network to provide
knowledge about related word pairs.  Lexical
cohesiveness between two words is calculated in the network
by ``activating'' the node for one word and observing the ``activity
value'' at the other word after some number of iterations of ``spreading
activation'' between nodes.  The network is trained automatically using a
language-specific knowledge source (a dictionary of definitions).
Kozima generalizes lexical cohesiveness to apply to a window of text,
and plots the cohesiveness of successive text windows in a document,
identifying the valleys in the measure as segment boundaries.  

A graphically motivated segmentation technique called {\it
dotplotting} is offered in  \cite{Reynar:94}.  This technique
uses a simplified notion of lexical cohesion, depending exclusively on
word repetition to find tight regions of topic similarity.

\subsection{Decision trees}

\cite{Litman:95} presents an algorithm that uses 
decision trees to combine multiple linguistic features extracted from
corpora of spoken text, including prosodic and lexical cues.  The
decision tree algorithm, like ours, chooses from a space of candidate
features, some of which are similar to our vocabulary questions.  The
set of candidate questions in Litman and Passonneu's approach, however,
is lacking in features related to ``lexical cohesion.'' In our work we
incorporate such features by using a pair of language models, as
described below.

\section{A Feature-Based Approach}
\label{sec:elm}
Our attack on the segmentation problem is based on a statistical framework that
we call {\it feature induction\/} for random fields and exponential models
\cite{Berger:96a,DellaPietra:96a}.  The idea is to construct a model which
assigns to each position in the data stream a probability that a boundary
belongs at that position.  This probability distribution arises by incrementally
building a log-linear model that weighs different ``features'' of the
data. For simplicity, we assume that the features are binary questions.

To illustrate (and to show that our approach is in no way restricted to text),
consider the task of partitioning a stream of multimedia data containing audio,
text and video. In this setting, the features might include 
questions such as:

{\nopagebreak
\medskip
{\small
\begin{itemize}
\item [$\bullet$]
{\em Does the phrase {\ww coming up} appear in the last utterance
of the decoded speech?}
\item [$\bullet$]
{\em Is there a sharp change in the video stream in the last $20$ frames?}
\item [$\bullet$]
{\em Does the language model degrade in performance in the next two
utterances?}
\item [$\bullet$]
{\em Is there a ``match'' between the spectrum of the current image and
an image near the last segment boundary?}
\item [$\bullet$]
{\em Are there blank video frames nearby?}
\item [$\bullet$]
{\em Is there a sharp change in the audio stream in the next utterance?}
\end{itemize}
}
\medskip
}

The idea of using features is a natural one, and indeed other recent
work on segmentation, such as \cite{Litman:95}, adopts this approach.
We take a unique approach to incorporating the information
inherent in various features, using the statistical framework of
exponential models to choose the best features and combine them in a
principled manner.

\subsection{A short-range model of language}
\label{sec:trigram}
Central to our approach to segmenting is a pair of tools: a short- and
long-range model of language. Monitoring the relative behavior of these two
models goes a long way towards helping our segmenter sniff out natural breaks
in the text. In this section and the next, we describe these language
models and explain their utility in identifying segments.

The trigram models $\triprob(w\given w_{-2}, w_{-1})$ we employ
use the Katz backoff scheme \cite{Katz:87a} for smoothing.
We trained trigram models on two different corpora.  The
Wall Street Journal corpus (\WSJ) is a $38$-million word corpus 
of articles from the newspaper.
The model was constructed using a set $\V$ of the approximately $20,000$ most
frequently occurring words in the corpus. 
Another model was constructed on the
Broadcast News corpus (\BN), made up of approximately 
$150$ million words (four and a half years) 
of transcripts of various news broadcasts, including  CNN news,
political roundtables, NPR broadcasts, and interviews.  

By restricting the conditioning information to the previous two words, the
trigram model is making the simplifying assumption---clearly false---that the
use of language one finds in television, radio, and newspaper can be modeled by
a second-order Markov process. Although words prior to $w_{-2}$ certainly bear
on the identity of $w$, higher-order models are impractical: the number of
parameters in an $n$-gram model is $O({\nV}^n)$, and finding the resources to
compute and store all these parameters becomes a hopeless task for
$n>3$. Usually the lexical myopia of the trigram model is a hindrance; however,
we will see how a segmenter can in fact make positive use of this
shortsightedness.

\subsection{A long-range model of language}
\label{sec:long-range}

One of the fundamental characteristics of language, viewed as a stochastic
process, is that it is highly {\it nonstationary\/}.  Throughout a written
document and during the course of spoken conversation, the topic evolves,
affecting local statistics on word occurrences. A model which could adapt to
its recent context would seem to offer much over a stationary model
such as the trigram model. For example, an adaptive model might, for some
period of time after seeing a word like {\ww homerun}, boost the probabilities
of the words {\ww \{homerun, pitcher, fielder, error, batter, triple,
out\}}. For an empirically-driven example, we provide an excerpt from the
\BN\ corpus.  Emphasized words mark where a long-range language model might
reasonably be expected to outperform (assign higher probabilities than) a
short-range model:

{\samepage{
\begin{quotation}
\noindent 
Some doctors are more {\bf skilled} at doing the {\bf procedure} than others
so it's {\bf recommended} that {\bf patients} ask {\bf doctors} about their
track record. People at high {\bf risk} of {\bf stroke} include those over age
55 with a family {\bf history} or high {\bf blood pressure}, {\bf diabetes} and
{\bf smokers}. We urge them to be evaluated by their family {\bf physicians}
and this can be done by a very simple {\bf procedure} simply by having them
{\bf test} with a {\bf stethoscope} for {\bf symptoms} of blockage.
\end{quotation}
}}

One means of injecting long-range awareness into a language model is by
retaining a cache of the most recently seen $n$-grams which is smoothed
together (typically by linear interpolation) with the static model; see for
example \cite{Jelinek:91a,Kuhn:90}.  Another approach, using maximum entropy
methods, introduces a parameter for {\it trigger pairs} of mutually informative
words, so that the occurrence of certain words in recent context boosts the
probability of the words that they trigger \cite{Lau:93}.

The method we use here, described in \cite{Beeferman:97a}, employs a static
trigram model as a ``prior,'' or default distribution, and adds certain
features to a family of conditional exponential models to capture some of the
nonstationary features of text.  The features are simple trigger pairs
of words chosen on the basis of mutual information.  Figure
\ref{tab:trigger-examples} provides a small sample of the $(s,t)$ trigger pairs
used in most of the experiments we will describe.

\begin{table}[ht]
\begin{center}
\small\ww
\begin{tabular}{| l | l |}
\hline
\multicolumn{1}{|c|}{$(s,t)$}& \multicolumn{1}{c|}{$e^\lambda$}\\
\hline
residues, carcinogens & 2.3\\
Charleston, shipyards & 4.0\\
microscopic, cuticle & 4.1\\
defense, defense & 8.4\\
tax, tax & 10.5\\
Kurds, Ankara & 14.8\\
Vladimir, Gennady & 19.6\\
Steve, Steve & 20.7\\
education, education & 22.2\\
music,  music & 22.4\\
insurance, insurance &  23.0\\
Pulitzer, prizewinning & 23.6\\
Yeltsin, Yeltsin & 23.7\\
Russian, Russian & 26.1\\
sauce, teaspoon & 27.1\\
flower, petals & 32.3\\
casinos, Harrah's & 42.8\\
drug, drug &  47.7\\
Claire, Claire & 80.9\\
picket, scab & 103.1\\
\hline
\end{tabular} 
\end{center}
\caption{A sample of the $84,694$ word pairs from the {\BN}
domain. Roughly speaking, after seeing an ``s'' word, the empirical probability
of witnessing the corresponding ``t'' in the next $\W$ words is boosted by the
factor in the third column. In the experiments described herein, $\W=500$. A
separate set of $(s,t)$ pairs were extracted from the \WSJ\ corpus.}
\label{tab:trigger-examples}
\end{table}

To incorporate triggers into a long-range language model, we begin by
constructing a standard, static backoff trigram model 
$\triprob(w\given w_{-2}, w_{-1})$ as described in 
\ref{sec:trigram}.  We then build a family of
conditional exponential models of the general form
\begin{eqnarray*}
\lefteqn{\expprob(w\given H) = \nonumber}\\
&& {1\over Z(H)} \exp\left({\sum_{i} \lambda_i f_i(H,w)}\right)
\, \triprob(w\given w_{-2}, w_{-1})
\label{eq:exp-form}
\end{eqnarray*}
where \mbox{$H \equiv w_{-\W}, w_{-\W+1}, \ldots, w_{-1}$} is the word {\it
history\/} (the $\W$ words preceding $w$ in the text), and $Z(H)$ is the
normalization constant
\begin{eqnarray*}
\lefteqn{Z(H) = } \\
&& \sum_{w\in\V} \exp\left({\sum_{i} \lambda_i f_i(H, w)}\right)
\triprob(w\given w_{-2}, w_{-1})\,.\nonumber
\end{eqnarray*}
The functions $f_i$, which depend both on the word history $H$ and the word
being predicted, are the features; each $f_i$ is assigned a weight $\lambda_i$.
In the models that we built, feature $f_i$ is an indicator function, testing
for the occurrence of a trigger pair $(s_i, t_i)$:
\begin{displaymath}
f_{i}(H, w) =  \cases{ 1 & if $s_i\in H$ and $w = t_i$\cr
                              0 & otherwise.} 
\end{displaymath}

The above equations reveal that the probability of a word $t$
involves a sum over all words $s$ such that $s\in H$ ($s$ appeared in the past
$500$ words) and $(s,t)$ is a trigger pair. One propitious manner of viewing
this model is to imagine that, when assigning probability to a word $w$
following a history of words $H$, the model ``consults'' a cache of words which
appeared in $H$ and which are the left half of some $(s,t)$ trigger pair.  In
general, the cache consists of content words $s$ which promote the probability
of their mate $t$, and correspondingly demote the probability of other
words. As described in \cite{Beeferman:97a}, for each $(s,t)$ trigger pair
there corresponds a real-valued parameter $\lambda$; the probability of $t$ is
boosted by a factor of $e^{\lambda}$ for $W$ words following the occurrence of
$s_i$.

The training algorithm we use for estimating the 
$\lambda$ values is the {\it
Improved Iterative Scaling\/} algorithm of 
\cite{DellaPietra:96a}, which is a scheme for
solving the maximum likelihood problem that is ``dual'' to
a corresponding maximum entropy problem.  Assuming
robust estimates for the $\lambda$ parameters, the resulting model is
essentially guaranteed to be superior to the trigram model.

For a concrete example, if \mbox{$s_i=${\ww Vladimir}} and $t_i=$\mbox{{\ww
Gennady}}, then $f_i=1$ if and only if {\ww Vladimir} appeared in the past $\W$
words and the current word $w$ is {\ww Gennady}. Consulting
Table~\ref{tab:trigger-examples}, we see that in the \BN\ corpus, the presence
of {\ww Vladimir} will boost the probability of {\ww Gennady} by a factor of
$19.6$ for the next $\W=500$ words.

\subsection{Language model ``relevance'' features} 
\def\relevance{R}
\def\contextrelevance{R_{C}}
\def\instrelevance{R}

A long-range language model such as that described in
Section~\ref{sec:long-range} 
uses selected words from the past ten, twenty or more
sentences to inform its decision on the possible identity of the next
word. This is likely to help if all of these sentences are in the
same document as the current word, for in that case the model has presumably
begun to adapt to the idiosyncracies of the current document. In the case of
the trigger model described above, the cache will be filled with ``relevant''
words.  In this setting, one would expect a long-range model to outperform a
trigram (or other short-range) model, which doesn't avail itself of long-range
information.

On the other hand, if the present document has just recently begun, the
long-range model is wrongly conditioning its decision on information
from a different---and presumably unrelated---document. A soap
commercial, for instance, doesn't benefit a long-range model in
assigning probabilities to the words in the news segment following the
commercial.  Often a long-range model will actually be misled by such
irrelevant context; in this case, the myopia of the trigram model is
actually helpful.

By monitoring the long- and short-range models, one might be more inclined
towards a partition when the long-range model suddenly shows a dip in
performance---a lower assigned probability to the observed words---compared to
the short-range model. Conversely, when the long-range model is consistently
assigning higher probabilities to the observed words, a partition is less
likely.

This motivates a quantitative measure of ``relevance,'' which we
define as the logarithm of the ratio of the probability the
exponential model assigns to the next word (or
sentence) to that assigned by the
short-range trigram model:
\begin{displaymath}
\instrelevance(H,w) \equiv 
   \log\left(\frac{\expprob(w\mid H)}{\triprob(w\mid w_{-2} w_{-1})}\right)\,.
\end{displaymath}
When the exponential model outperforms the trigram model,
$\instrelevance>0$.  

\begin{figure}[t]
\centerline{\hskip-20pt
\psfig{figure=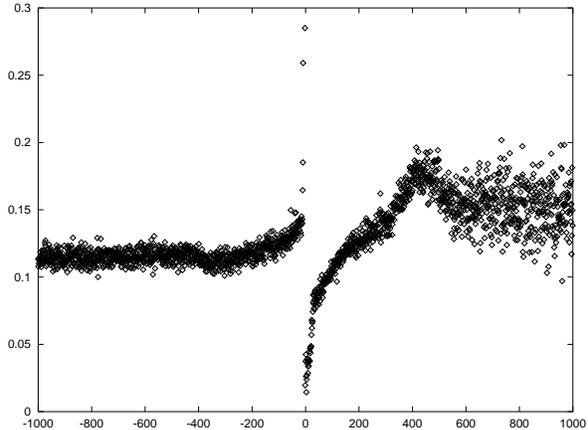,width=3.2in,angle=-90}}
\caption{Near the beginning of a segment, an
adaptive, long-range language model is on average less accurate than 
a static trigram model.  The figure plots
the average value of the logarithm 
of the ratio of the adaptive language model to the
static trigram model as a function of relative position
in the segment, with position zero indicating the beginning of
a segment.  The statistics were collected over the roughly seven million
words of mixed broadcast news and Reuters data comprising the \TDT\
corpus (see Section~\ref{sec:construction}).}
\label{fig:instrel}
\end{figure}

If we observe the behavior of $\instrelevance$ as a
function of the position of the word within a segment, we find that on
average $\instrelevance$ slowly increases from below zero to well above
zero. Figure~\ref{fig:instrel} gives a striking graphical
illustration of this phenomenon.  
The figure plots
the average value of $R$ as a function of relative position
in the segment, with position zero indicating the beginning of
a segment.
This plot shows that 
when a segment boundary is crossed
the predictions of the adaptive model undergo a dramatic
and sudden degradation, and then steadily become more accurate as relevant
content words for the new segment are encountered and added to
the cache.  (The few very high points to the left of a segment boundary
are primarily a consequence of the word {\ww cnn}---which is a
trigger word and often appears at the beginning and end of
a broadcast news segment.)

This observed behavior is consistent with our earlier intuition: the
cache of the long-range model is destructive early in a document, when
the new content words bear little in common with the content words from
the previous article.  Gradually, as the cache fills with words drawn
from the current article, the long-range model gains steam and
$\instrelevance$ improves.  While Figure~\ref{fig:instrel} shows that
this behavior is very pronounced as a ``law of large numbers,'' our
feature induction results indicate that relevance is also a very good
predictor of boundaries for individual events.

In the experiments we report in this paper, 
we assume that sentence boundaries are
provided in the annotation, and so the questions we ask are actually about
the relevance score assigned to entire sentences normalized by
sentence length, a geometric mean of language model ratios.

\long\def\ignore#1{}

\ignore{
Raw instantaneous relevance is a useful feature for segmentation, but
it can be significantly improved.  In particular, there
are certain trigram contexts in which the trigger feature cache is
likely to pay off, and certain contexts in which the trigram model
alone provides all of the necessary predictive information.  Often
this distinction is a domain-specific matter of style.  For instance,
in the {\it Wall Street Journal} an occurrence of {\ww mr.} or {\ww mrs.}
often indicates that a proper name triggered earlier in the document
will follow.  Similarly, {\ww shareholders of...} predicts a company
name that has already been introduced, and {\ww los} predicts the second
half of a Spanish city name tied closely with the topic.  Other
contexts are tightly bound (with low entropy) to a small set of
possible successor words in a way that is independent of the
long-range history.  Computing the ``context relevance'', the expected
value of the instantaneous relevance for a given context, for all one-
and two-word contexts in the training data gives us a ranking
excerpted in Table \ref{tab:contextrel}.  For bigram contexts $(w_{-2},
w_{-1})$, the context-relevance $\contextrelevance(w_{-2}, w_{-1})$ is as follows:
\begin{equation}
\contextrelevance(w_{-2}, w_{-1}) = 
\sum_{w \in \V, H=w_{-N},\dots, w_{-2}, w_{-1}}
\widetilde{p}(H, w)
\log\left(\frac{\expprob(w | H)}
{\triprob(w | w_{-2}, w_{-1})}\right)\,,
\end{equation}
where $\widetilde{p}$ is the empirical probability
distribution.

\begin{table}
\begin{tt}
\begin{center}
\begin{tabular}{|r|l|} 
\hline
        {\ww mr.}       &     1.402986 \\
        {\ww ms.}       &     1.134188 \\
        {\ww mrs.}      &     1.063191 \\
        {\ww doctor}    &     0.874050 \\
        {\ww winter}    &     0.799196 \\
        {\ww el}        &     0.707947 \\
        {\ww senator}   &     0.696035 \\[2pt]
	$\vdots\quad$   &     $\quad\quad\vdots$ \\[2pt]
       {\ww five}       &    -0.525790 \\
       {\ww freeman}    &    -0.556112 \\
       {\ww eighteen}   &    -0.569623 \\
       {\ww zero}       &    -0.600722 \\
       {\ww fairchild}  &    -0.614652 \\
       {\ww row}        &    -0.688912 \\
       {\ww billion}    &    -0.722252 \\
\hline
\end{tabular}
\label{tab:contextrel}
\caption{The top and bottom few unigram contexts in terms of average
instantaneous relevance in the training data.  Note that the top contexts
are {\it introductory} in nature.}
\end{center}
\end{tt}
\end{table}

Normalizing instantaneous relevance by the context-dependent
expectation above gives us a final relevance indicator $\relevance_i$ (where
$i$ is a position in the corpus) that we can incorporate into our
segmentation model.}

\subsection{Vocabulary features} 

In addition to the estimate of ``topicality'' that relevance features provide,
we included features pertaining to the identity of words before and after
potential segment boundaries as candidates in our exponential model.  The set
of candidate word-based features we use are simple questions of the
form 

{\nopagebreak
{\small\it 
\begin{itemize}
\item [$\bullet$] 
Does the word appear up to 1 sentence in the future?  2 sentences? 3? 5? 
\item [$\bullet$] 
Does the word appear up to 1 sentence in the past?  2 sentences ? 3? 5? 
\item [$\bullet$] 
Does the word appear up to 5 sentences in the past but not 5 sentences in the future? 
\item [$\bullet$] 
Does the word appear up to 5 sentences in the future but not 5 sentences in the past?
\item [$\bullet$] 
Does the word appear up to 1 word in the future?  5 words? 
\item [$\bullet$] 
Does the word appear up to 1 word in the past?  5 words? 
\item [$\bullet$] 
Does the word begin the preceding sentence? 
\end{itemize}
}
}

\section{Feature Induction} 
\label{sec:fi}
\def\qz{q_{\mbox{\tiny 0}}}
\def\div{\,\raise.1ex\hbox{$\|$}\,}
\def\Q{{\cal Q}}
\def\C{{\cal C}}
\def\reals{{\bf R}}
\def\gain{{G}}

To cast the problem of determining segment boundaries in statistical terms, we
set as our goal the construction of a probability distribution $q(b\given
\omega)$, where $b\in\{\yes,\no\}$ is a random variable describing the presence
of a segment boundary in context $\omega$.  We consider distributions in the
{\it linear exponential family} $\Q(f,\qz)$ given by
\begin{displaymath}
\Q(f, \qz) = \left\{ 
  q(b\given\omega) = {1\over Z_\lambda(\omega)}
  e^{\lambda\cdot f(\omega)} \; \qz(b\given \omega)\right\}
\end{displaymath}
where $\qz(b\given \omega)$ is a prior or {\it default}
distribution on the presence of a boundary, and $\lambda\cdot f(\omega)$
is a linear combination of binary features $f_i(\omega)\in\{0,1\}$ with
real-valued {\it feature parameters\/} $\lambda_i$:
\begin{displaymath}
\lambda\cdot f(\omega) = \lambda_1 f_1(\omega) + \lambda_2 f_2(\omega) +
\cdots \lambda_n f_n(\omega)\,.
\end{displaymath}
The normalization
constants
\begin{displaymath}
Z_\lambda(\omega) = 1+ e^{\lambda\cdot f(\omega)}
\end{displaymath}
insure that this is indeed a family of conditional probability distributions.
(This family of models is closely related to the class of sigmoidal
belief networks \cite{Neal:92}.)

Our judgment of the merit of a model \mbox{$q\in \Q(f,\qz)$} relative to a
reference distribution $p\not\in \Q(f,\qz)$ during training is made in terms of
the Kullback-Leibler divergence
\begin{displaymath}
D(p\div q) = \sum_{\omega\in \Omega} p(\omega) \sum_{b\in\{\syes,\sno\}}
p(b\given\omega) \log {p(b\given \omega) \over q(b\given \omega)}\,.
\end{displaymath}
Thus, when $p$ is chosen to be the empirical distribution of a sample of
training events $\{(\omega, b)\}$, we are using the maximum likelihood
criterion for model selection.  Under certain mild regularity conditions, the
maximum likelihood solution
\begin{displaymath}
q^\star = \argmin_{q\in\Q(f,\qz)} D(p\div q)
\end{displaymath}
exists and is unique.  To find this solution, we use the iterative
scaling algorithm presented in \cite{DellaPietra:96a}.  

This explains how a model is chosen once we know the
features $f_1, \ldots, f_n$, but how are these features to be found?
The procedure that we follow is a greedy algorithm akin to growing
a decision tree.  Given an initial distribution $q$ and a set of candidate
features $\C$, we consider  the one-parameter family
of distributions $\{ q_{\alpha, g}\}_{\alpha\in\reals} = \Q(g, q)$
for each $g\in\C$.
The {\it gain} of the candidate feature $g$ is defined to be
\begin{displaymath}
\gain_q(g) = \argmax_{\alpha} \left(D(\pt\div q) - D(\pt\div q_{\alpha, f})\right)\,.
\end{displaymath}
This is the improvement to the model that would result from
adding the feature $g$ and adjusting its weight to the best value.
After calculating the gain of each candidate feature, the one with the
largest gain is chosen to be added to the model, and all of the model's
parameters are then adjusted using iterative scaling.  In this manner,
an exponential model is incrementally built up using the most
informative features.

\medskip

\begin{figure}[ht]
\centerline{\psfig{figure=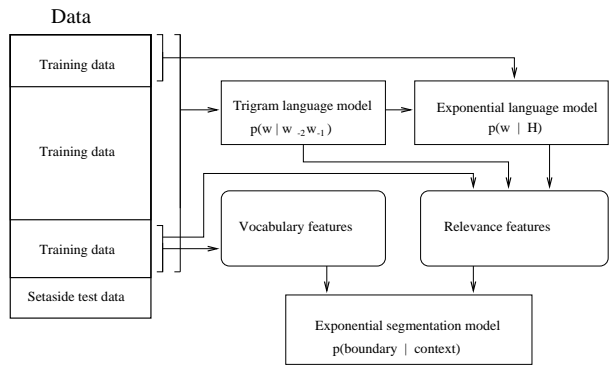,width=3.15in}}
\caption{Data flow in training the exponential segmentation model}
\label{fig:dataflow}
\end{figure}

Having concluded our discussion of our overall approach,
we present in Figure~\ref{fig:dataflow} a schematic
view of the steps involved in building a segmenter using
this approach.

\section{Feature Induction in Action}
\label{sec:construction}
\def\rshift{-290}
\def\leftone{-160}
\def\lefttwo{-120}
\def\leftthree{-80}
\def\leftfour{-40}
\def\leftfive{0}
\def\vshift{37}  
\def\sentbars{
 \begin{picture}(0,\vshift)(\rshift,0)
 \put(0,-5){\line(0,1){6}}
 \put(0,18){\makebox(0,0){\shortstack{\small\sc current \\ \small\sc position}}}
 \put(130,-5){\line(0,1){6}}
 \put(130,10){\makebox(0,0){\shortstack{+1}}}
 \put(170,-5){\line(0,1){6}}
 \put(170,10){\makebox(0,0){\shortstack{+2}}}
 \put(210,-5){\line(0,1){6}}
 \put(210,10){\makebox(0,0){\shortstack{+3}}}
 \put(250,-5){\line(0,1){6}}
 \put(250,10){\makebox(0,0){\shortstack{+4}}}
 \put(290,-5){\line(0,1){6}}
 \put(290,10){\makebox(0,0){\shortstack{+5}}}
 \put(-130,-5){\line(0,1){6}}
 \put(-130,10){\makebox(0,0){\shortstack{-1}}}
 \put(-170,-5){\line(0,1){6}}
 \put(-170,10){\makebox(0,0){\shortstack{-2}}}
 \put(-210,-5){\line(0,1){6}}
 \put(-210,10){\makebox(0,0){\shortstack{-3}}}
 \put(-250,-5){\line(0,1){6}}
 \put(-250,10){\makebox(0,0){\shortstack{-4}}}
 \put(-290,-5){\line(0,1){6}}
 \put(-290,10){\makebox(0,0){\shortstack{-5}}}
 \put(-290,-3){\line(1,0){580}}
 \end{picture}
 \vskip0pt
}
\def\wordone#1#2{
 \begin{picture}(0,\vshift)(\rshift,0)
 \put(0,0){\circle*{5}}
 \put(25,0){\vector(-1,0){25}}
 \put(25,0){\makebox(80,0){\shortstack{\small\sc #1\\ $#2$}}}
 \put(105,0){\vector(1,0){25}}
 \put(130,0){\circle*{5}}
 \end{picture}
}
\def\sentone#1#2{
 \begin{picture}(0,\vshift)(\rshift,0)
 \put(0,-3){\line(0,1){6}}
 \put(25,0){\vector(-1,0){25}}
 \put(25,0){\makebox(80,0){\shortstack{\small\sc #1\\ $#2$}}}
 \put(105,0){\vector(1,0){25}}
 \put(130,-3){\line(0,1){6}}
 \end{picture}
}
\def\sentmone#1#2{
 \begin{picture}(0,\vshift)(\leftone,0)
 \put(0,-3){\line(0,1){6}}
 \put(25,0){\vector(-1,0){25}}
 \put(25,0){\makebox(80,0){\shortstack{\small\sc #1\\ $#2$}}}
 \put(105,0){\vector(1,0){25}}
 \put(130,-3){\line(0,1){6}}
 \end{picture}
}
\def\senttwo#1#2{
 \begin{picture}(0,\vshift)(\rshift,0)
 \put(0,-3){\line(0,1){6}}
 \put(45,0){\vector(-1,0){45}}
 \put(45,0){\makebox(80,0){\shortstack{\small\sc #1\\ $#2$}}}
 \put(125,0){\vector(1,0){45}}
 \put(170,-3){\line(0,1){6}}
 \end{picture}
}
\def\sentmtwo#1#2{
 \begin{picture}(0,\vshift)(\lefttwo,0)
 \put(0,-3){\line(0,1){6}}
 \put(45,0){\vector(-1,0){45}}
 \put(45,0){\makebox(80,0){\shortstack{\small\sc #1\\ $#2$}}}
 \put(125,0){\vector(1,0){45}}
 \put(170,-3){\line(0,1){6}}
 \end{picture}
}
\def\sentthree#1#2{
 \begin{picture}(0,\vshift)(\rshift,0)
 \put(0,-3){\line(0,1){6}}
 \put(65,0){\vector(-1,0){65}}
 \put(65,0){\makebox(80,0){\shortstack{\small\sc #1\\ $#2$}}}
 \put(145,0){\vector(1,0){65}}
 \put(210,-3){\line(0,1){6}}
 \end{picture}
}
\def\sentmthree#1#2{
 \begin{picture}(0,\vshift)(\leftthree,0)
 \put(0,-3){\line(0,1){6}}
 \put(65,0){\vector(-1,0){65}}
 \put(65,0){\makebox(80,0){\shortstack{\small\sc #1\\ $#2$}}}
 \put(145,0){\vector(1,0){65}}
 \put(210,-3){\line(0,1){6}}
 \end{picture}
}
\def\sentfour#1#2{
 \begin{picture}(0,\vshift)(\rshift,0)
 \put(0,-3){\line(0,1){6}}
 \put(85,0){\vector(-1,0){85}}
 \put(85,0){\makebox(80,0){\shortstack{\small\sc #1\\ $#2$}}}
 \put(165,0){\vector(1,0){85}}
 \put(250,-3){\line(0,1){6}}
 \end{picture}
}
\def\sentmfour#1#2{
 \begin{picture}(0,\vshift)(\leftfour,0)
 \put(0,-3){\line(0,1){6}}
 \put(85,0){\vector(-1,0){85}}
 \put(85,0){\makebox(80,0){\shortstack{\small\sc #1\\ $#2$}}}
 \put(165,0){\vector(1,0){85}}
 \put(250,-3){\line(0,1){6}}
 \end{picture}
}
\def\sentfive#1#2{
 \begin{picture}(0,\vshift)(\rshift,0)
 \put(0,-3){\line(0,1){6}}
 \put(105,0){\vector(-1,0){105}}
 \put(105,0){\makebox(80,0){\shortstack{\small\sc #1\\ $#2$}}}
 \put(185,0){\vector(1,0){105}}
 \put(290,-3){\line(0,1){6}}
 \end{picture}
}
\def\sentmfive#1#2{
 \begin{picture}(0,\vshift)(\leftfive,0)
 \put(0,-3){\line(0,1){6}}
 \put(105,0){\vector(-1,0){105}}
 \put(105,0){\makebox(80,0){\shortstack{\small\sc #1\\ $#2$}}}
 \put(185,0){\vector(1,0){105}}
 \put(290,-3){\line(0,1){6}}
 \end{picture}
}
\def\wordfive#1#2{
 \begin{picture}(0,\vshift)(\rshift,0)
 \put(0,0){\circle*{5}}
 \put(105,0){\line(-1,0){105}}
 \put(105,0){\makebox(80,0){\shortstack{\small\sc #1\\ $#2$}}}
 \put(185,0){\line(1,0){105}}
 \put(290,0){\circle*{5}}
 \end{picture}
}
\def\wordmfive#1#2{
 \begin{picture}(0,\vshift)(\leftfive,0)
 \put(0,0){\circle*{5}}
 \put(105,0){\line(-1,0){105}}
 \put(105,0){\makebox(80,0){\shortstack{\small\sc #1\\ $#2$}}}
 \put(185,0){\line(1,0){105}}
 \put(290,0){\circle*{5}}
 \end{picture}
}
\def\beginsmone#1#2{
 \begin{picture}(0,\vshift)(\leftone,0)
 \put(0,0){\makebox(50,0){\shortstack{\small\sc #1\\ $#2$}}}
 \put(0,0){\circle*{5}}
 \end{picture}
}
\def\sentfig{
 \begin{picture}(35,0)(0,-3)
 \put(0,-3){\line(0,1){6}}
 \put(10,0){\vector(-1,0){10}}
 \put(10,0){\makebox(10,0){}}
 \put(20,0){\vector(1,0){10}}
 \put(30,-3){\line(0,1){6}}
 \end{picture}
}
\def\wordfig{
 \begin{picture}(35,0)(0,-3)
 \put(0,0){\circle*{4}}
 \put(10,0){\line(-1,0){10}}
 \put(10,0){\makebox(10,0){}}
 \put(20,0){\line(1,0){10}}
 \put(30,0){\circle*{4}}
 \end{picture}
}
\def\sentoneex#1#2{
 \begin{picture}(70,15)(0,-3)
 \put(0,-3){\line(0,1){6}}
 \put(20,0){\vector(-1,0){20}}
 \put(20,0){\makebox(30,0){\shortstack{\small\sc #1\\ $#2$}}}
 \put(50,0){\vector(1,0){20}}
 \put(70,-3){\line(0,1){6}}
 \put(70, 8){\makebox(0,0){\small +1}}
 \end{picture}
}
\def\wordfiveex#1#2{
 \begin{picture}(70,15)(0,-3)
 \put(0,0){\circle*{4}}
 \put(20,0){\line(-1,0){20}}
 \put(20,0){\makebox(30,0){\shortstack{\small\sc #1\\ $#2$}}}
 \put(50,0){\line(1,0){20}}
 \put(70,0){\circle*{4}}
 \put(70, 8){\makebox(0,0){\small +5}}
 \end{picture}
}
\def\compoundfiveex#1#2{
 \begin{picture}(140,15)(0,-3)
 \put(0,-3){\line(0,1){6}}
 \put(0, 8){\makebox(0,0){\small -5}}
 \put(20,0){\vector(-1,0){20}}
 \put(20,0){\makebox(30,0){\shortstack{\small\sc #1}}}
 \put(50,0){\vector(1,0){20}}
 \put(70,-3){\line(0,1){6}}
 \put(70,-3){\line(0,1){6}}
 \put(90,0){\vector(-1,0){20}}
 \put(90,0){\makebox(30,0){\shortstack{\small\sc $\neg$ #1\\ $#2$}}}
 \put(120,0){\vector(1,0){20}}
 \put(140,-3){\line(0,1){6}}
 \put(140, 8){\makebox(0,0){\small +5}}
 \end{picture}
}
\long\def\ignore#1{}

This section provides a peek at the 
construction of segmenters for two different
domains.  Inspecting the sequence of
features selected by the induction algorithm reveals much about feature
induction in general, and how it applies to the segmenting task in particular.
We emphasize that the process of feature selection is completely automatic once
the set of candidate features has been selected.

The first segmenter was built on the \WSJ\ corpus.
The second was built on the Topic Detection and Tracking Corpus
\cite{Allan:97}.  The 
\TDT\ corpus is a mixed collection of newswire articles and broadcast 
news transcripts adapted from text corpora previously released by the
Linguistic Data Consortium; in particular, portions of data were
extracted from the 1995 and 1996 Language Model text collections
published by the LDC in support of the DARPA Continuous Speech
Recognition project.  The extracts used for \TDT\ include material from
the Reuters newswire service, and from the Primary Source Media CD-ROM
publications of transcripts for news programs that appeared on the ABC,
CNN, NPR and PBS broadcast networks; the size of the corpus is roughly
7.5 million words.  The \TDT\ corpus was constructed as part of a
DARPA-sponsored project intended to study methods for detecting new
topics or events and tracking their reappearance and evolution over
time.

\subsection{\WSJ\ features}

For the \WSJ\ experiments, which we describe first, a total of $300,000$
candidate features were available to the induction program.  Though the trigram
prior was trained on $38$ million words, the trigger parameters were
only trained on a one million word subset of this data.

\setlength{\unitlength}{.2ex}

Figure~\ref{fig:wsjfeats} shows the first several features that were
selected by the feature induction algorithm.  This shows the word or
relevance score for each feature together with the value of $e^\lambda$
for the feature after iterative scaling is complete for the final model.
The \sentfig figures indicate features that are active over a range of
sentences.  Thus, the symbol
\sentoneex{\ww mr.}{0.07} represents the feature ``Does the word {\ww
mr.} appear in the next sentence?'' which, if true, contributes 
a factor of $e^\lambda=0.07$ to the exponential model.  Similarly,
the \wordfig figures represent features that are active over
a range of words.  For example, the figure
\wordfiveex{\ww he}{0.08} represents the question ``Does the word
{\ww he } appear in the next five words?'' which is assigned a
weight of  $0.08$.  The symbol
\compoundfiveex{said}{2.7} stands for a feature which asks
``Does the word {\ww said} appear in the previous five sentences but {\it not}
in the next five sentences?'' and contributes a factor of $2.7$ if the answer
is ``yes.''

\def\hs{\hskip0in}
\def\vs{\vskip0pt}
\begin{figure}[ht]
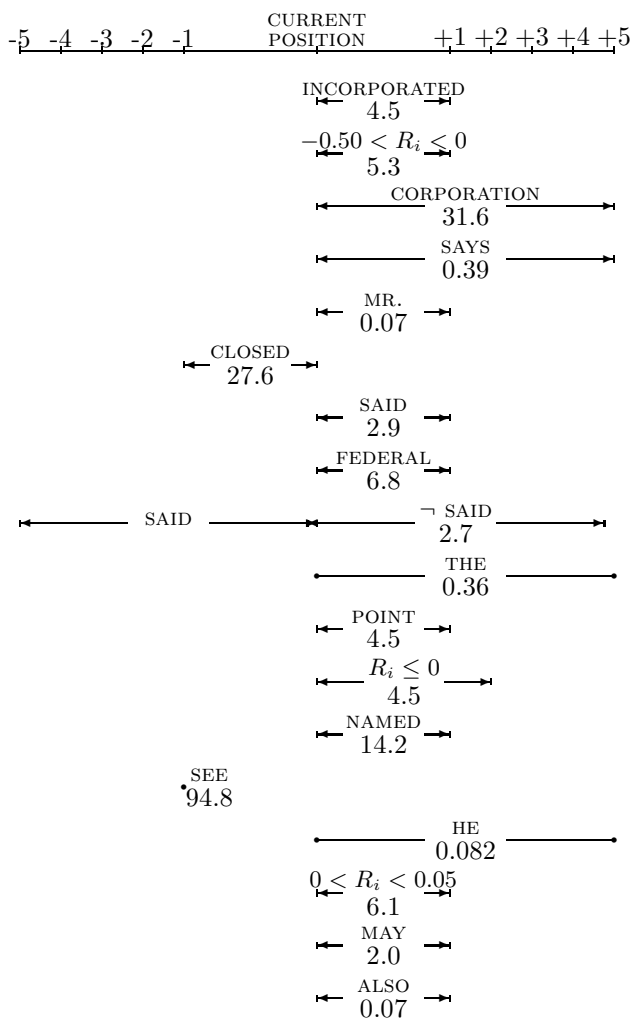

\setlength{\unitlength}{.09ex}
\def\vshift{49}  
\hs\sentbars
\hs\sentone{incorporated}{4.5} \vs
\hs\sentone{$-0.50 < R_i < 0$}{5.3} \vs
\hs\sentfive{corporation}{31.6} \vs
\hs\sentfive{says}{0.39} \vs
\hs\sentone{mr.}{0.07} \vs
\hs\sentmone{closed}{27.6} \vs
\hs\sentone{said}{2.9} \vs
\hs\sentone{federal}{6.8} \vs
\hs\sentmfive{said}{}\hskip-7pt\sentfive{$\neg$ said}{2.7} \vs
\hs\wordfive{the}{0.36} \vs
\hs\sentone{point}{4.5} \vs
\hs\senttwo{$R_i \leq 0$}{4.5} \vs
\hs\sentone{named}{14.2} \vs
\hs\beginsmone{see}{94.8} \vs
\hs\wordfive{he}{0.082} \vs
\hs\sentone{$0<R_i<0.05$}{6.1} \vs
\hs\sentone{may}{2.0} \vs
\hs\sentone{also}{0.07} \vs
\mbox{}
\caption{First several features induced for the \WSJ\ corpus, presented
in order of selection, with $e^\lambda$ factors underneath.  The length
of the bars indicate active range of the feature, in words or
sentences, relative to the current word.}
\label{fig:wsjfeats}
\end{figure}

Most of the features in Figure~\ref{fig:wsjfeats} make a good deal of
sense. The first selected feature, for instance, is a strong hint that an
article may have just begun; articles in the \WSJ\ corpus often concern
companies, and typically the full name of the company ({\ww Acme Incorporated},
for instance) only appears once at the beginning of the article, 
and subsequently in abbreviated form ({\ww Acme}). 
Thus the appearance of {\ww incorporated} is
a strong indication that a new article may have recently begun.

The second feature uses the relevance statistic\footnote{For the \WSJ\
experiments, we modified the language model relevance statistic
by adding a weight to each word position depending only on its
trigram history $w_{-2}, w_{-1}$.  Although our results require
further analysis, we do not believe that this makes a significant
difference in the features chosen by the algorithm, or the
quantitative performance of the resulting segmenter.}.
If the trigger model performs
poorly relative to the trigram model in the following sentence, this feature
(roughly speaking) boosts the probability of a segment at this location by
a factor of $5.3$. 

The fifth feature concerns the presence of the word {\ww Mr.}  In
hindsight, we can explain this feature by noting that in \WSJ\ data the
style is to introduce a person in the beginning of an article by
writing, for example, {\ww Wile E. Coyote, president of Acme
Incorporated...}  and then later in the article using a shortened form
of the name: {\ww Mr. Coyote cited a lack of explosives...}  Thus, the
presence of {\ww Mr.} in the following sentence {\it discounts\/} the
probability of an article boundary by $0.07$, a factor of roughly $14$.

The sixth feature---which boosts the probability of a segment if the previous
sentence contained the word {\ww closed}---is another artifact of the
\WSJ\ domain, where articles often end with a statement of a company's
performance on the stock market during the day of the story of interest.
Similarly, the end of an article is often made with an invitation to
visit a related story; hence a sentence beginning with {\ww see} boosts
the probability of a segment boundary by a large factor of $94.8$.
Since a personal pronoun typically requires an antecedent, the presence
of {\ww he} among the first words is a sign that the current position is
{\it not} near an article boundary, and this feature therefore has a
discounting factor of $0.082$.

\subsection{\TDT\ features}

For the \TDT\ experiments, a larger vocabulary and 
roughly 800,000 candidate features were
available to the induction program.  Though the trigram prior was
trained on approximately $150$ million words, the trigger parameters
were trained on a 10 million word subset of the \BN\ corpus.

\begin{figure}[ht]
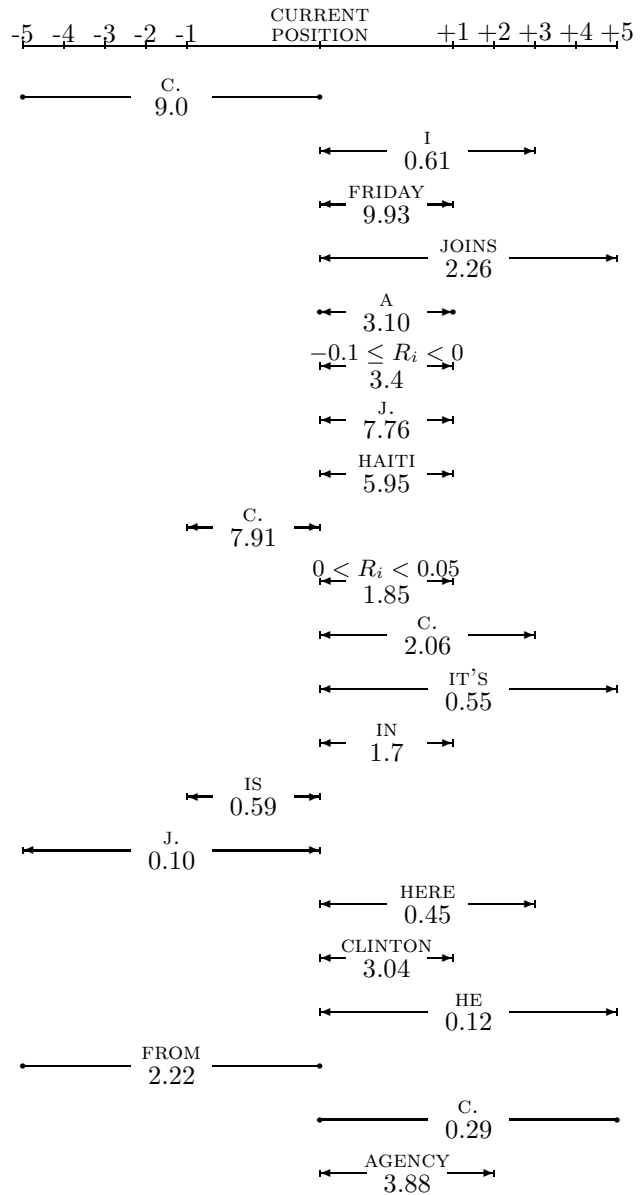

\setlength{\unitlength}{.09ex}
\def\vshift{50}
\hs\sentbars
\hs\wordmfive{c.}{9.0} \vs
\hs\sentthree{i}{0.61} \vs
\hs\sentone{friday}{9.93} \vs
\hs\sentfive{joins}{2.26} \vs
\hs\wordone{a}{3.10} \vs
\hs\sentone{$-0.1 \leq R_i < 0$}{3.4} \vs
\hs\sentone{j.}{7.76} \vs
\hs\sentone{haiti}{5.95} \vs
\hs\sentmone{c.}{7.91} \vs
\hs\sentone{$0<R_i<0.05$}{1.85} \vs
\hs\sentthree{c.}{2.06} \vs
\hs\sentfive{it's}{0.55} \vs
\hs\sentone{in}{1.7} \vs
\hs\sentmone{is}{0.59} \vs
\hs\sentmfive{j.}{0.10} \vs
\hs\sentthree{here}{0.45} \vs
\hs\sentone{clinton}{3.04} \vs
\hs\sentfive{he}{0.12} \vs
\hs\wordmfive{from}{2.22} \vs
\hs\wordfive{c.}{0.29} \vs
\hs\senttwo{agency}{3.88} \vs
\mbox{}
\caption{First several features induced for the {\bf TDT}\ corpus, presented
in order of selection, with $e^\lambda$ factors underneath.}
\label{fig:tdtfeats}
\end{figure}

Figure~\ref{fig:tdtfeats} reveals the first several features chosen by
the induction algorithm.  The letter {\ww c.} appears among several of
the first features.  This is because of the fact that the data is
tokenized for speech processing (whence {\ww c. n. n.} rather than {\ww
cnn}), and the network identification information is often given at the
end and beginning of news segments ({\ww c. n. n.'s richard blystone is
here to tell us...}).  The first feature asks if the letter {\ww c.}
appears in the previous five words; if so, the probability of a segment
boundary is boosted by a factor of $9.0$.  The personal pronoun {\ww i}
appears as the second feature; if this word appears in the following
three sentences then the probability of a segment boundary is
discounted.  

The language model relevance statistic appears for the first time in the
sixth feature.  The word {\ww j.} that the seventh and fifteenth
features ask about can be attributed to the large number of news stories
in the data having to do with the O.J. Simpson trial.  The nineteenth
feature asks if the term {\ww from} appears among the previous
five words, and if the answer is ``yes'' raises the probability of 
a segment boundary by more than a factor of two.  This feature
makes sense in light of the ``sign-off'' conventions that news
reporters and anchors follow ({\ww This is Wolf Blitzer
reporting live from the White House}).
Similar explanations of many of the remaining features are
easy to guess from a perusal of Figure~\ref{fig:tdtfeats}.

\section{A Probabilistic Error Metric}
\label{sec:error}
Precision and recall statistics are commonly used in natural
language processing and information
retrieval to assess the quality of algorithms.  For the segmentation
task they might be used to gauge how frequently boundaries actually occur
when they are hypothesized and vice versa.  Although they have snuck
into the literature in this disguise, we believe they are unwelcome
guests.

A useful error metric should somehow correlate with the utility of the
instrumented procedure in a real application. In almost any conceivable
application, a segmenting tool that consistently comes close---off by a
sentence, say---is preferable to one that places boundaries willy-nilly. Yet
an algorithm that places a boundary a sentence away from the actual
boundary every time actually receives {\em worse} precision and recall scores
than an algorithm that hypothesizes a boundary at every position.  It
is natural to expect that in a segmenter, close should count for something.

A useful metric should also be robust with respect to the scale
(words, sentences, paragraphs, for instance) at which boundaries are
determined. However, precision and recall are scale-dependent
quantities. \cite{Reynar:94} uses an error window that redefines
``correct'' to mean hypothesized within some constant window of units
away from a reference boundary, but this approach still suffers from
overdiscretizing error, drawing all-or-nothing lines insensitive to
gradations of correctness.

Finally, for many purposes it is useful to have a metric that 
is a single number. A commonly cited flaw of
the precision/recall figures is their complementary nature: hypothesizing more
boundaries raises precision at the expense of recall, allowing an algorithm
designer to tweak parameters to trade precision for recall. One proposed
work-around is to employ dynamic time warping to come up with an explicit
alignment between the segments proposed by the algorithm and the reference
segments, and then to combine insertion, deletion, and substitution errors into
an overall penalty.  This error metric, in common use in speech recognition,
can be achieved by a similar Viterbi search.  
A string edit distance such as this 
is useful and reasonable for applications like speech or
spelling correction partly because it measures how much work a
user would have to do to correct the output of the machine.  
For many of the applications we envision for segmentation, however, 
the user will not correct the output but will rather browse 
the returned text to extract information.  

Our proposed metric satisfies the listed desiderata. It formalizes in a
probabilistic manner the effect of document co-occurrence on goodness, in which
it is deemed desirable for related units of information to appear in the same
document and unrelated units to appear in separate documents.

\subsection{The new metric}
\def\D{D}
\def\P{P}

Segmentation, whether at the word or sentence level, is about
identifying boundaries between successive units of information in a
text corpus.  Two such units are either related or unrelated by the
intent of the document author.  A natural way to reason about
developing a segmentation algorithm is therefore to optimize the
likelihood that two such units are correctly labeled as being related
or being unrelated.  Our error metric $\P_\mu$ is simply the {\it
probability that two sentences drawn randomly from the corpus are
correctly identified as belonging to the same document or not
belonging to the same document}.  More formally, given
two segmentations {\tt ref} and {\tt hyp} for a corpus $n$
sentences long, 
\begin{displaymath}
\P_\mu(\mbox{\small\tt ref},\mbox{\small\tt hyp}) 
= \sum_{1\leq i\leq j\leq n} \hskip-5pt \D_\mu(i,j) \;
\delta_{\mbox{\small\tt ref}}(i,j)  \ \overline{\oplus}\
\delta_{\mbox{\small\tt hyp}}(i,j)
\end{displaymath}
Here $\delta_{\mbox{\small\tt ref}}$ is an indicator function which is
1 if the two corpus indices specified by its parameters belong in the
same document, and 0 otherwise; similarly, $\delta_{\mbox{\small\tt
hyp}}$ is 1 if the two indices are hypothesized to belong in the same
document, and 0 otherwise.  The $\overline{\oplus}$ operator is the {\tt
XNOR} function (``both or neither'') on its two operands.  The function
$\D_\mu$ is a {\it distance probability distribution} over the set of
possible distances between sentences chosen randomly from the corpus,
and will in general depend on certain parameters $\mu$ such as the
average spacing between sentences.  If $\D_\mu$ is uniform over the
length of the text, then the metric represents the probability that any
two sentences drawn from the corpus are correctly identified as being in
the same document or not.

Consider the implications of this for information retrieval.  Suppose
there is precisely one sentence in a target corpus that satisfies our
information demands.  For some applications it may be sufficient for
the system to return only that sentence, but in general we desire that
it return as many sentences directly related to the target sentence as
possible, without returning too many unrelated sentences. If we assume
``related'' to mean ``contained in the same document'', then our error
metric judges algorithms based on how often this happens.

In practice letting $\D_\mu$ be the uniform distribution is
unreasonable, since for large corpora most randomly drawn pairs of
sentences are in different documents and are correctly identified as
such by even the most naive algorithms.  We instead adopt a
distribution that focuses on small distances.  In particular, we choose
$\D_\mu$ to be an exponential distribution with mean $1/\mu$, a parameter
that we fix at the approximate mean document length for the domain:
\begin{displaymath}
\D_\mu(i,j) = \gamma_\mu\, e^{-\mu|i-j|}\,.
\end{displaymath}
In the above, $\gamma_\mu$ is a normalization chosen so that $\D_\mu$ is a
probability distribution over the range of distances it can accept.

There are several sanity checks that validate the use of our metric.
The measure is a probability and therefore a real number between 0 and
1.  We expect 1 to represent perfection; indeed, an algorithm scores 1
with respect to some data if and only if it predicts its segmentation
exactly.  It captures the notion of nearness in a principled way, gently
penalizing algorithms that hypothesize boundaries that aren't quite
right, and scaling down with the algorithm's degradation.  Furthermore,
it is not possible to ``cheat'' and obtain a high score with this
metric: spurious behavior such as never hypothesizing boundaries and
hypothesizing nothing but boundaries are penalized.  We refer to
Section~\ref{sec:results} for sample results on how these 
trivial algorithms score.

One weakness of the metric as we have presented it here is that there is no
principled way of specifying the distance distribution $\D_\mu$.  We
plan to give a more detailed analysis of this problem and present a
method for choosing the parameters $\mu$ in a future paper.

\section{Experimental Results}  
\label{sec:results}

\subsection{Quantitative results}

\begin{figure*}[ht]
\begin{center}
\renewcommand{\arraystretch}{1.4}
\hskip-20pt\begin{tabular}{|c|c|c|c|c|c|c|}
\hline
\shortstack{\it model} 
       & \shortstack{ \it reference \\  \it segments} 
       & \shortstack{ \it hypoth.\raise10pt\hbox{\ } \\  \it segments} 
       & \shortstack{$ \quad \P_\mu \quad $} 
       & \shortstack{ \it precision}
       & \shortstack{ \it recall}
       & \shortstack{ \it F-measure}\\
\hline
{ \bf feature induction\raise10pt\hbox{\ }} &
757  &    792  &   83\% &   56\% &   54\% &  55  \\ \hline
\shortstack{ \bf random} &  
757  &    757  &   67\% &   17\% &   16\% &  17  \\ \hline
\shortstack{ \bf all} &  
757  &   13540 &   53\% &    5\% &  100\% &  10  \\ \hline
\shortstack{ \bf none} &  
757  &     0   &   52\% &    0\% &    0\% &  --- \\ \hline
\shortstack{ \bf even} &  
757  &    753  &   68\% &   17\% &   17\% &  17  \\ \hline
\end{tabular}
\end{center}
\medskip
\stepcounter{figure}
\label{wsjtable}
{Table \thefigure: Quantitative results for \WSJ\ segmentation.
The \WSJ\ model was trained on 325K words of data,
and tested on a similarly sized portion of unseen text.
The top 70 features were selected.  
The mean segment length in the training and test
data was $1/\mu=18$ sentences.
As a basis of comparison,
the figures for several baseline models are given.
The figures in the {\bf random} row were calculated 
by randomly generating a number of segments equal to the
number appearing in the test data.  The {\bf all} and {\bf none}
rows include the figures for models which hypothesize all
possible segment boundaries and no boundaries, respectively.
The {\bf even} row shows the results of simply hypothesizing
a segment boundary every 18 sentences.}

\vskip.3in
\begin{center}
\renewcommand{\arraystretch}{1.4}
\hskip-20pt\begin{tabular}{|c|c|c|c|c|c|c|}
\hline
\shortstack{\it model} 
       & \shortstack{ \it reference \\  \it segments} 
       & \shortstack{ \it hypoth.\raise10pt\hbox{\ } \\  \it segments} 
       & \shortstack{$ \quad \P_\mu \quad $} 
       & \shortstack{ \it precision}
       & \shortstack{ \it recall}
       & \shortstack{ \it F-measure}\\
\hline
\shortstack{ \bf feature induction\raise10pt\hbox{\ }
\\ {\small\bf (Model B)}} &  
9984 & 9543 & 88\% & 60\% & 57\% & 58 \\ \hline
\shortstack{ \bf feature induction\raise10pt\hbox{\ }
\\ {\small\bf (Model A)}}&  
9984 & 9449 & 82\% & 47\% & 45\% & 46 \\ \hline
\shortstack{ \bf random} &  
9984 & 9984 & 68\% & 12\% & 12\% & 12 \\ \hline
\shortstack{ \bf all} &  
9984 & 219,099 & 59\% & 5\% & 100\% & 9 \\ \hline
\shortstack{ \bf none} &  
9984 & 0 & 43\% & 0\% & 0\% & --- \\ \hline
\shortstack{ \bf even} &  
9984 & 9980 & 74\% & 14\% & 12\% & 13 \\ \hline
\end{tabular}
\end{center}
\medskip
\stepcounter{figure}
\label{tdttable}
{Table \thefigure: Quantitative results for \TDT\ segmentation.
The \TDT\ models were trained on 2M words and
tested on  4.3M words of previously  unseen \TDT\ data.
Model A was trained on 2M words of broadcast news
data from 1992--1993, not included
in \TDT\ corpus, and the top 100 features were selected.
Model B was trained on the first 2M words of TDT corpus which 
is made up of a mix of CNN transcripts and Reuters newswire,
and again the top 100 features were selected.
The mean document length was $1/\mu=25$ sentences.}
\end{figure*}

After feature induction was carried out (as described in
Section~\ref{sec:construction}), a simple decision procedure was used
for actually placing boundaries: a segment boundary was placed at each
position for which the model probability was above a fixed threshold
$\alpha$, with boundaries required to be separated by a minimum number
of sentences $\epsilon$.  The threshold and minimum separation were
determined on heldout data in order to maximize the probability
$\P_\mu$, and turned out to be $\alpha=0.20$ and $\epsilon=2$ for the
\WSJ\ model, and $\alpha=0.14$ and $\epsilon=5$ for the \TDT\ models.

The quantiative results for the \WSJ\ and \TDT\ models are collected
in Tables~5 and~6 respectively.  For the \WSJ\ model, the
probabilistic metric $P_\mu$ was $0.83$ when 
evaluated on 325K words of test data, 
and the precision and recall for {\it exact} matches
of boundaries were 56\% and 54\%, for an F-measure of 55.  As
a simple baseline we compared this performance to that obtained
by four simple default methods for assigning boundaries:
choosing boundaries randomly, assigning every possible 
boundary, assigning no boundaries, 
and deterministically placing a segment boundary every
$1/\mu$ sentences.   It is instructive to compare the values
of $\P_\mu$ with precision and recall for these default
algorithms in order to obtain some intuition for the new
error metric.  

Two separate models were built to segment the \TDT\ corpus.
The first, which we shall refer to simply as Model A, was trained
using two million words from the \BN\ corpus 
from the 1992-1993 time period.  This data contains CNN
transcripts, but no Reuters newswire data.  Model B was
trained on the first two million words of the \TDT\ corpus.
Both models were tested on the last 4.3 million words of
the \TDT\ corpus.   We expect Model A to be inferior to
Model B for two reasons: the lack of Reuters data in it's
training set and the difference of between one and two years
in the dates of the stories in the training and test sets.
The difference is quantifiied in Table~6, which
shows that $\P_\mu=0.82$ for Model~A while $\P_\mu=0.88$ for
Model~B.

\subsection{Qualitative results}

\begin{figure}[ht]
\begin{center}
\begin{tabular}{c}
\hskip-12pt\psfig{file=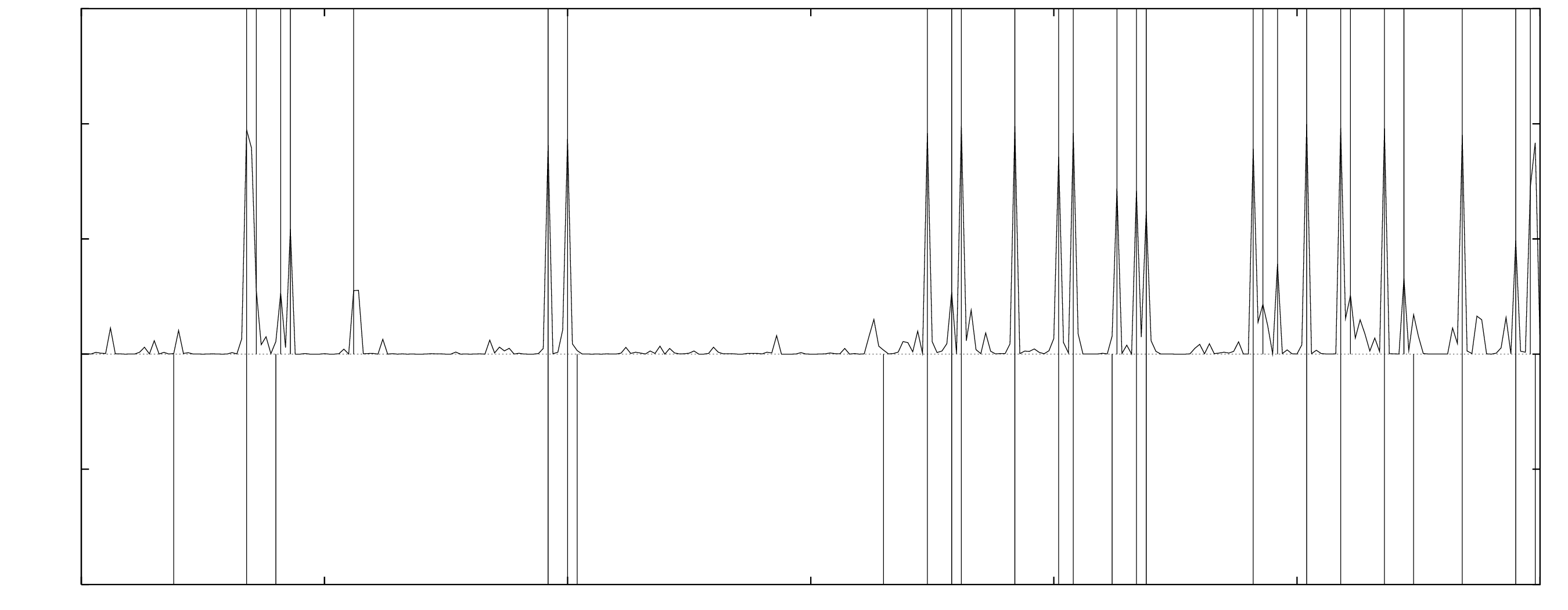,width=3.10in,height=1.1in,angle=-90} \\[8pt]
\hskip-12pt\psfig{file=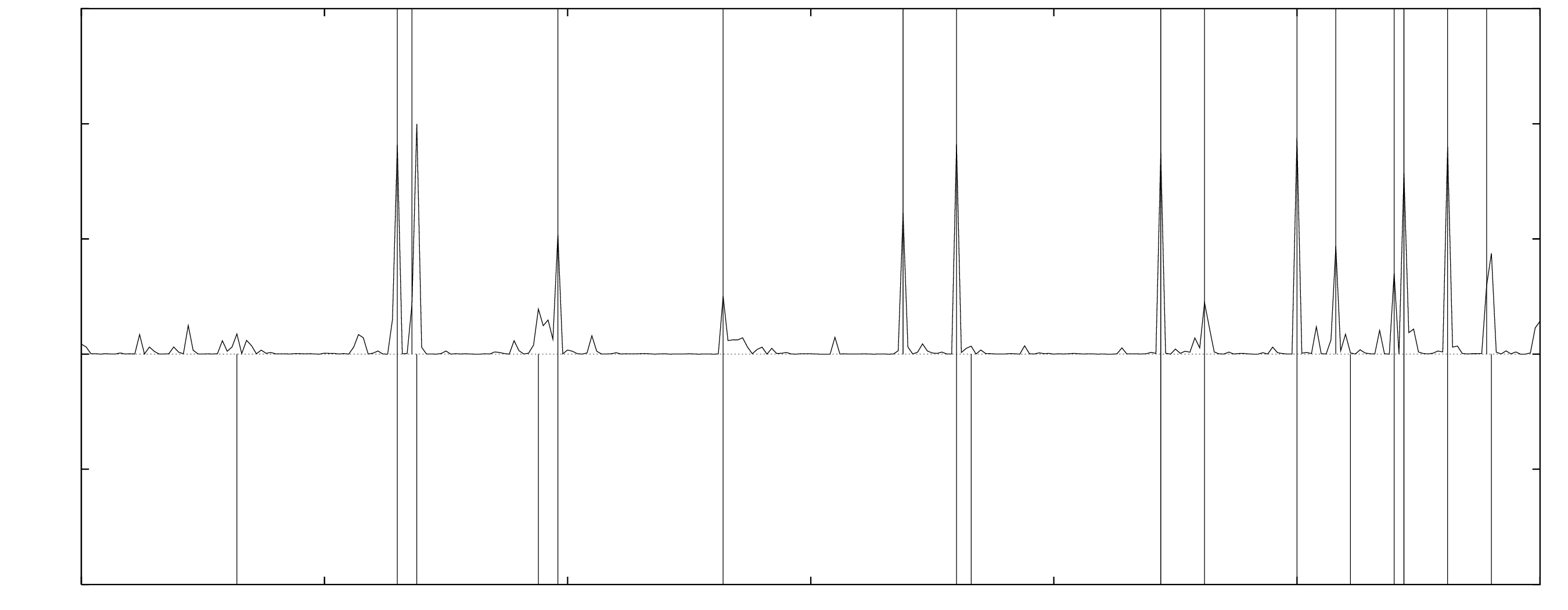,width=3.10in,height=1.1in,angle=-90} \\[8pt]
\hskip-12pt\psfig{file=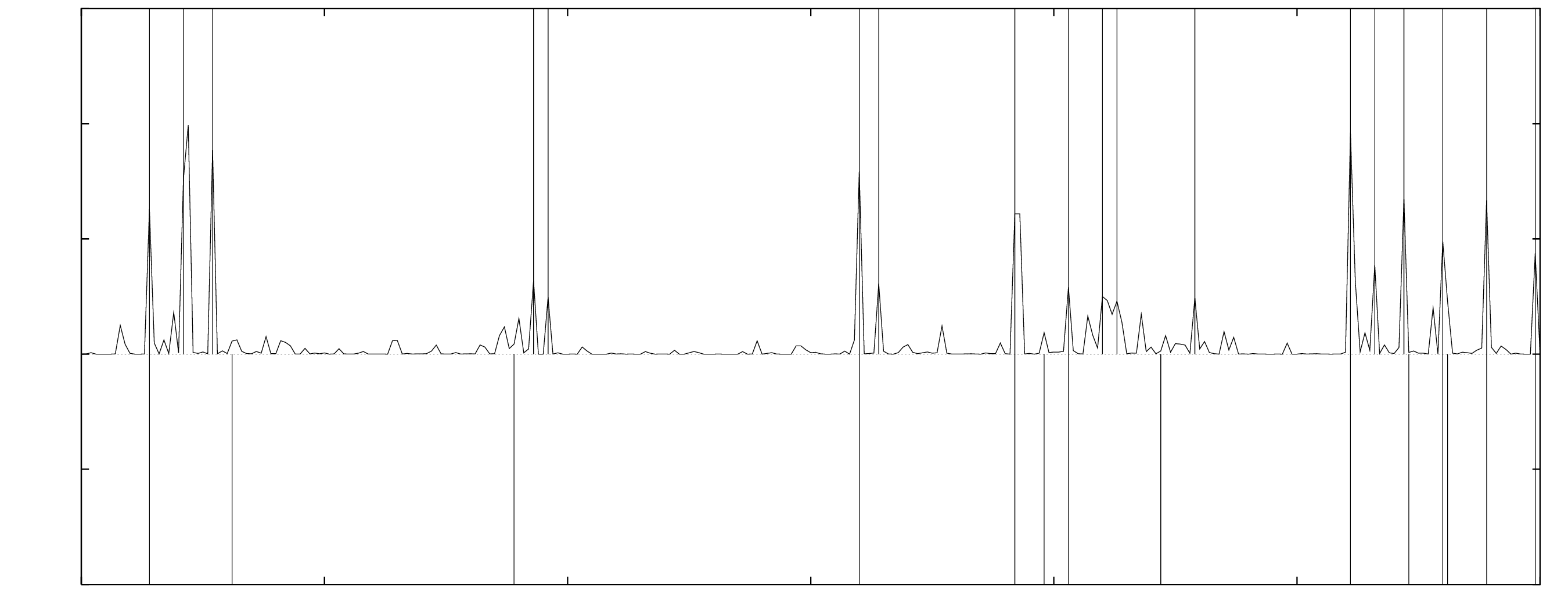,width=3.10in,height=1.1in,angle=-90} \\[8pt]
\hskip-12pt\psfig{file=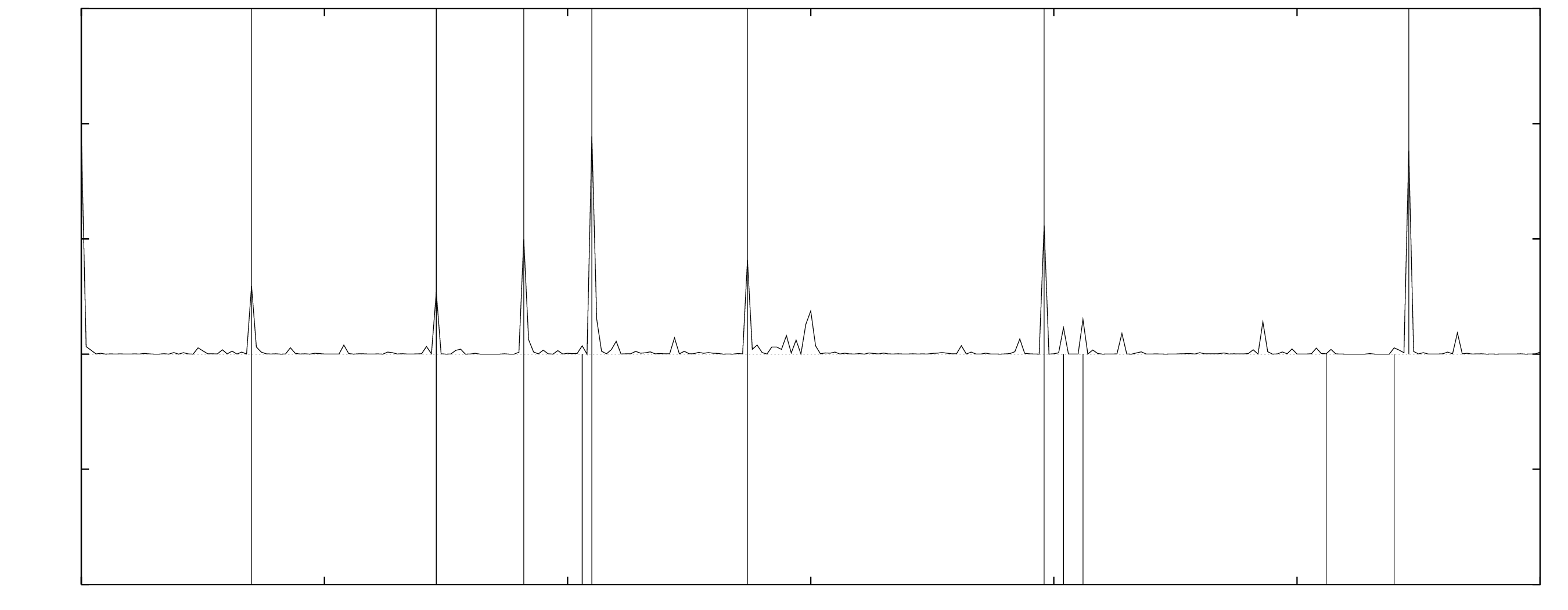,width=3.10in,height=1.1in,angle=-90} \\[8pt]
\hskip-12pt\psfig{file=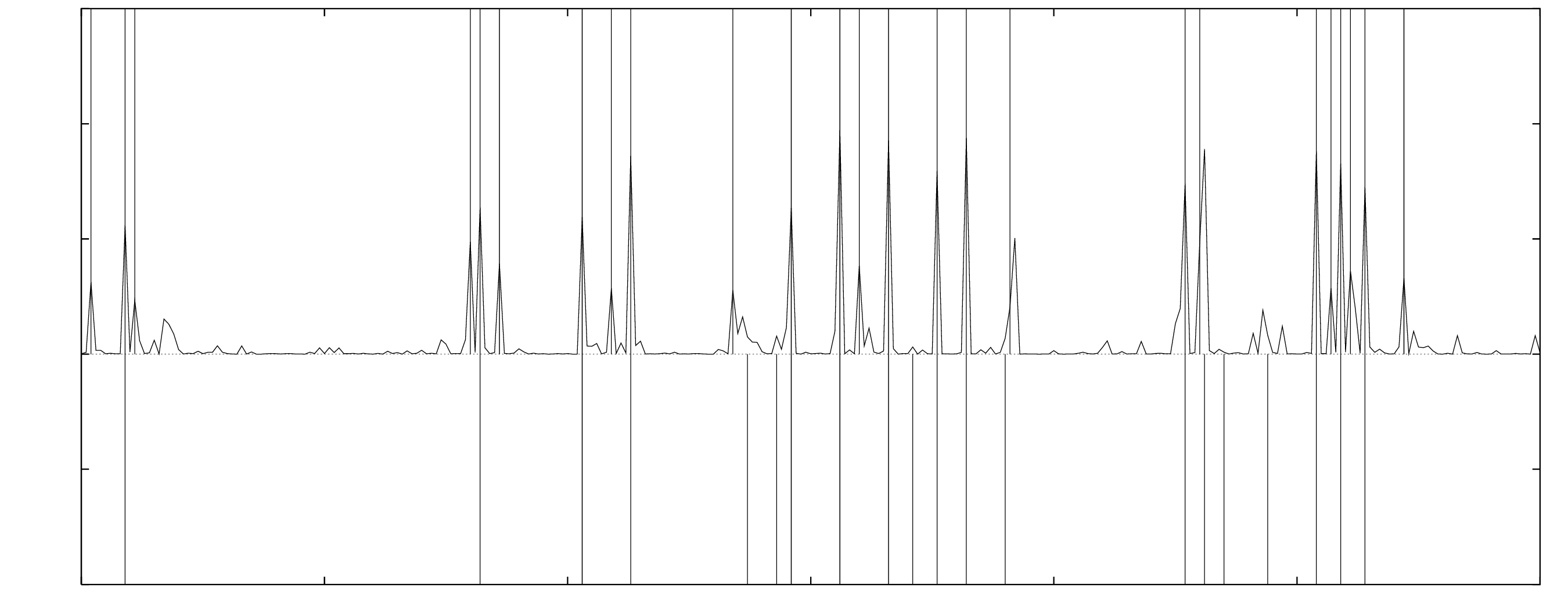,width=3.10in,height=1.1in,angle=-90} \\[8pt]
\end{tabular}
\end{center}
\caption{Typical segmentations of \WSJ\ test data.
The lower verticle lines indicate reference
segmentations (``truth'').  The upper verticle lines are
boundaries placed by the algorithm.  The fluctuating
curve is the probability of a segment boundary according
to the exponential model after $70$ features were induced.}
\label{fig:sample2}
\end{figure}

\begin{figure}[ht]
\begin{center}
\begin{tabular}{c}
\hskip-12pt\psfig{file=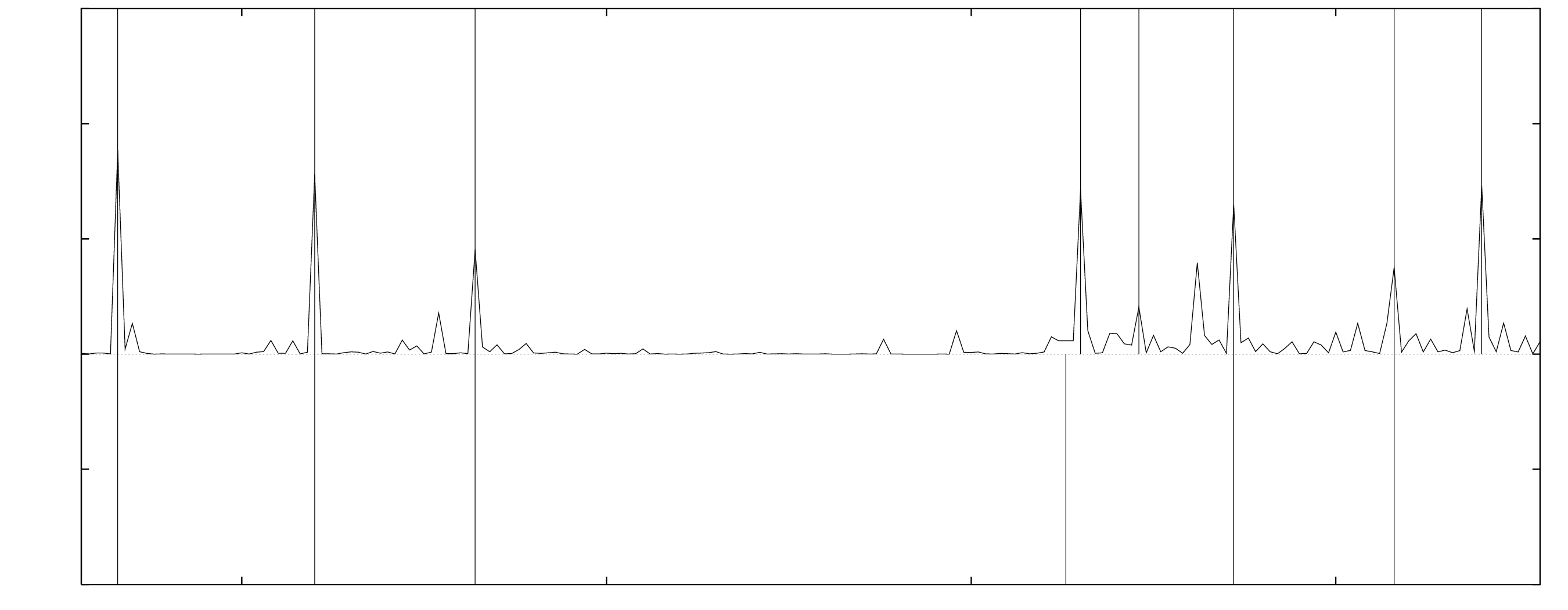,width=3.10in,height=1.1in,angle=-90} \\[8pt]
\hskip-12pt\psfig{file=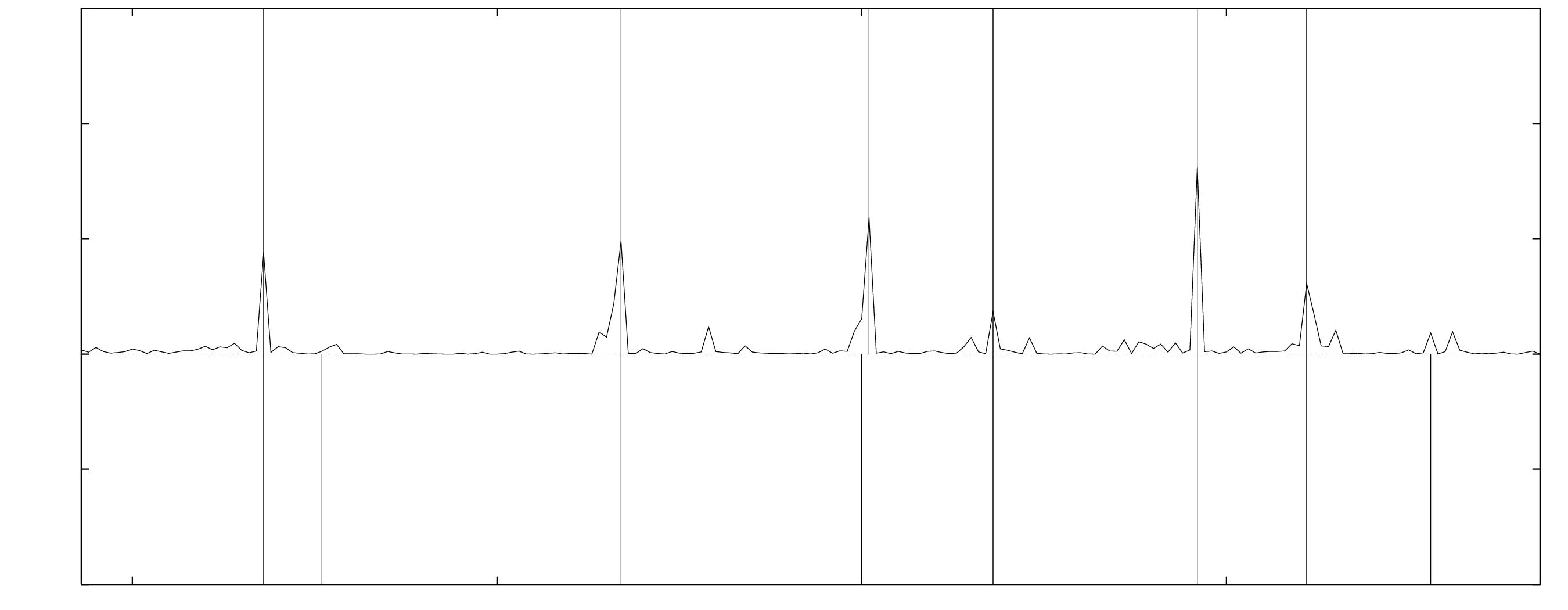,width=3.10in,height=1.1in,angle=-90} \\[8pt]
\hskip-12pt\psfig{file=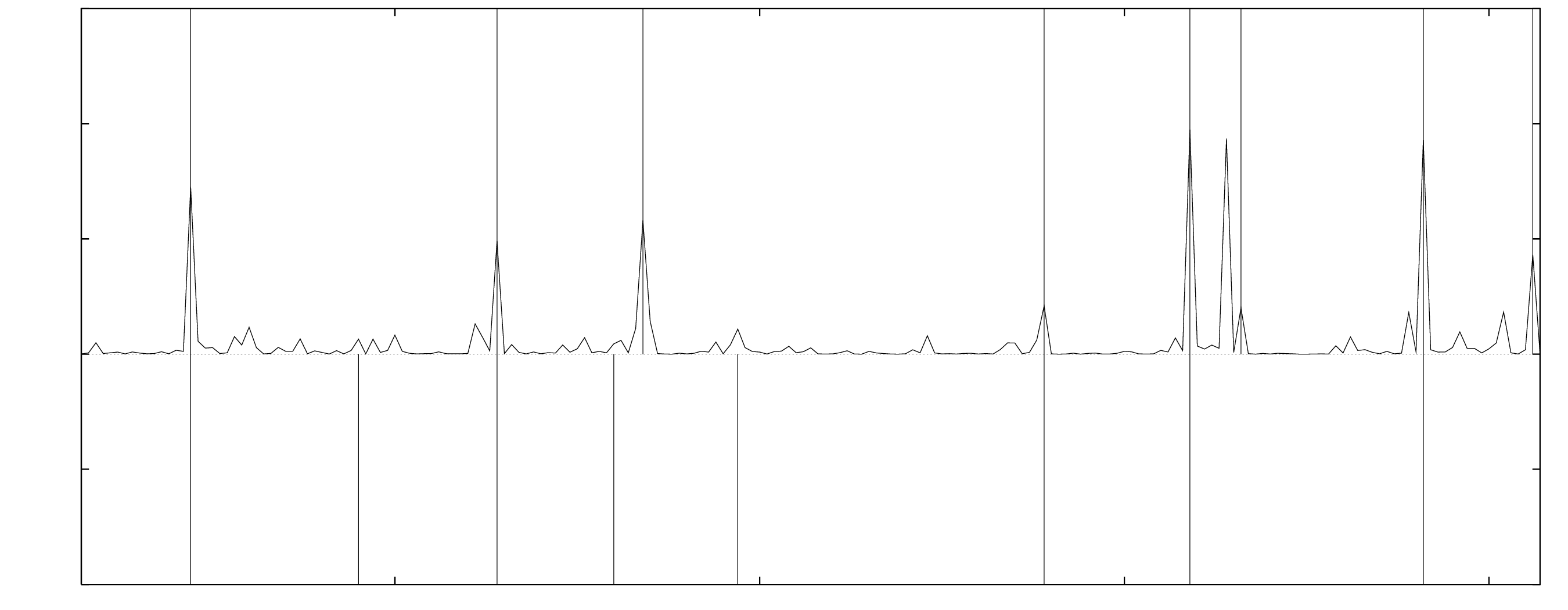,width=3.10in,height=1.1in,angle=-90} \\[8pt]
\hskip-12pt\psfig{file=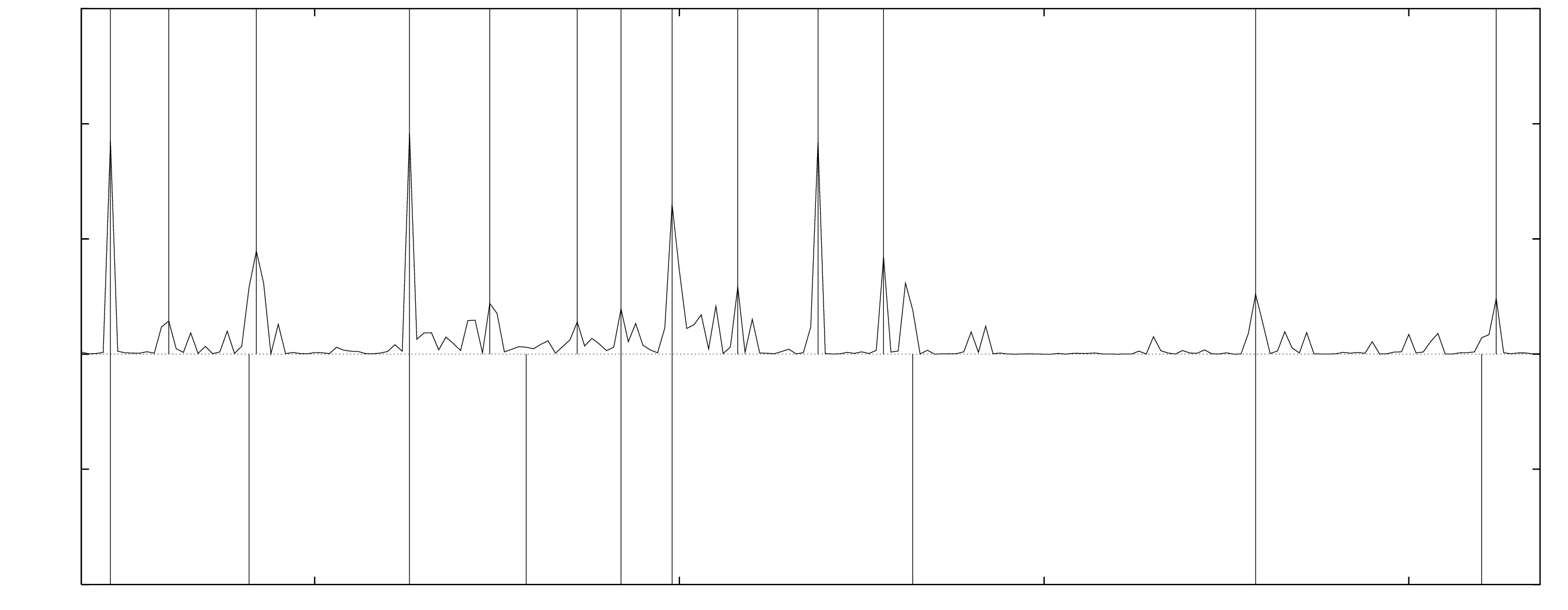,width=3.10in,height=1.1in,angle=-90} \\[8pt]
\hskip-12pt\psfig{file=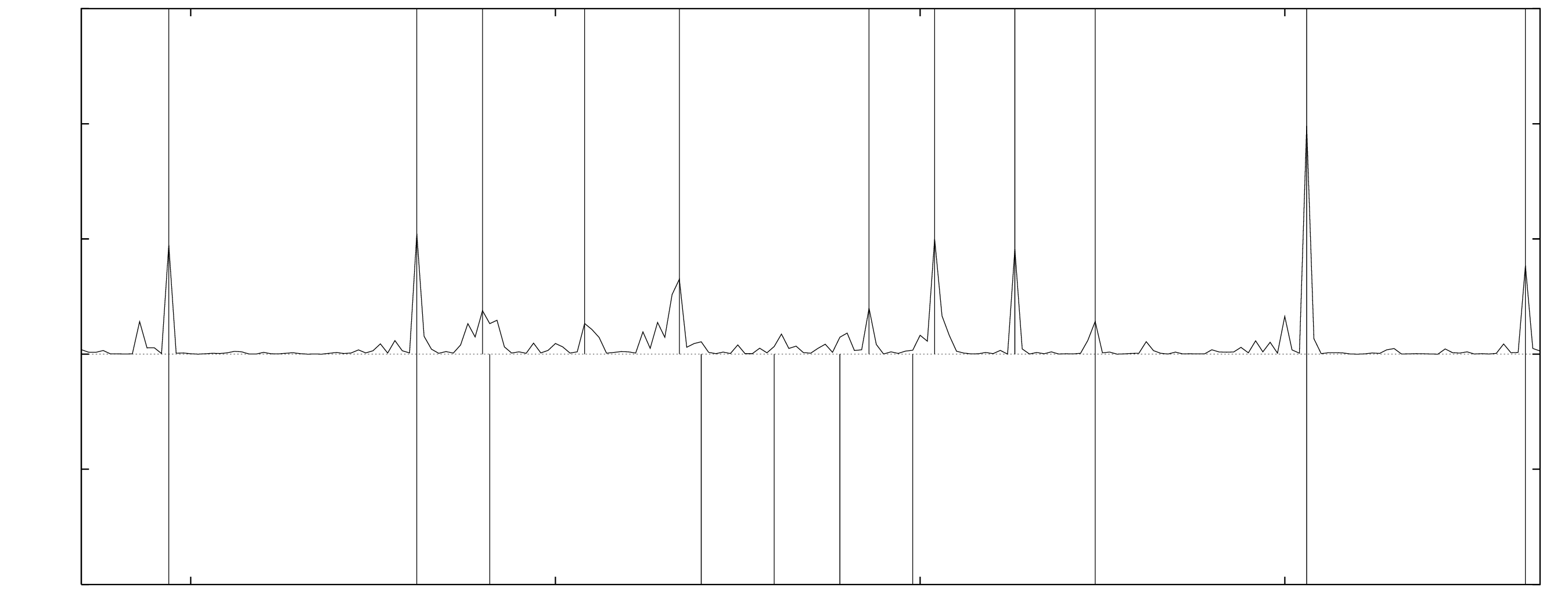,width=3.10in,height=1.1in,angle=-90}
\\[8pt]
\end{tabular}
\end{center}
\caption{Randomly chosen segmentations of \TDT\ test data,
in 200 sentence blocks, using Model B.}
\label{fig:tdtsample}
\end{figure}

We now present graphical examples of the segmentation algorithm at work
on previously unseen test data.  Figure~\ref{fig:sample2} shows the
performance of the \WSJ\ segmenter on a typical collection of test data, in
blocks of $300$ contiguous sentences.  In these figures the reference
segmentation is shown {\it below} the horizontal line as a vertical line
at the position between sentences where the article boundary occurred.
The decision made by the automatic segmenter is shown as a verticle line
{\it above} the horzontal line at the appropriate position.  The
fluctuating curve is the probability assigned by the exponential model
constructed using feature induction.  Notice that in this domain many of
the segments are quite short, adding special difficulties for the
segmentation problem.  Figure~\ref{fig:tdtsample} shows the performance
of the \TDT\ segmenter (Model B) on five randomly chosen blocks of 200
sentences from the \TDT\ test data.

We hasten to add that these results were obtained with no smoothing or
pruning of any kind, and with no more than $100$ features induced from
the candidate set of several hundred thousand.  Unlike many other
machine learning methods, feature induction for exponential models is
quite robust to overfitting since the features act in concert to assign
probability to events rather than splitting the event space and
assigning probability using relative counts.  We expect that significantly
better results can be obtained by simply training on much more
data, and by allowing a more sophisticated set of features.

\section{Conclusions} 
We have presented and evaluated a new statistical model for
segmenting unpartitioned text into coherent fragments.  We leverage
long- and short-range language models, as well as automatic feature
induction techniques, in the design of this model.  In this work we
rely exclusively on simple lexical features, including a topicality measure called
{\it relevance} and a number of vocabulary features that are induced
from a large space of candidate features.

We have proposed a new probabilistically motivated error metric for
the assessment of segmentation algorithms.  Qualitative assessment
as well as the evaluation of our algorithm
with this new metric demonstrates its
effectiveness in two very different domains, {\it Wall Street Journal}
articles and broadcast news transcripts.

Our immediate application of this model will be to the video-on-demand
application called {\it Informedia\/} \cite{Christel:95}.  We intend to
mix simple audio and video features such as statistics from pauses,
black frames, and color histograms with our lexical features in order to
segment news broadcasts into component stories.  Other applications that
we have not explored in this paper include automatic inference of
subtopic structure for information retrieval, document summarization,
and improved language modeling.

\section*{Acknowledgements}
We thank Michael Witbrock and Alex Hauptmann for discussions on the
segmentation problem within the context of the {\it Informedia\/}
project.  We also thank Jaime Carbonell and Yiming Yang for their input,
and for encouraging us to build segmentation models on the \TDT\ corpus.
Participants in the \TDT\ pilot study, including James Allan,
Rich Schwartz, Jon Yamron, and especially George Doddington, provided
invaluable feedback on the probabilistic evaluation metric.

\bibliography{segment}

\begin{thebibliography}{}

\bibitem[\protect\citename{Allan}{to~appear}]{Allan:97} 
Allan, J.
\newblock To appear.  
\newblock Topic Detection and Tracking Corpus,
\newblock Linguistic Data Consortium, University of Pennsylvania.

\bibitem[\protect\citename{Beeferman, Berger, and Lafferty}1997]{Beeferman:97a}
Beeferman, D., A.~Berger, and J.~Lafferty.
\newblock 1997.
\newblock A model of lexical attraction and repulsion.
\newblock In Proceedings of the 35th Annual Meeting of the ACL,
Madrid, Spain.

\bibitem[\protect\citename{Berger, {Della~Pietra}, and
  {Della~Pietra}}1996]{Berger:96a}
Berger, A., S.~{Della~Pietra}, and V.~{Della~Pietra}.
\newblock 1996.
\newblock A maximum entropy approach to natural language processing.
\newblock {\em Computational Linguistics}, 22(1):39--71.

\bibitem[\protect\citename{Christel \bgroup et al.\egroup }1995]{Christel:95}
Christel, M., T.~Kanade, M.~Mauldin, R.~Reddy, M.~Sirbu, S.~Stevens, and
  H.~Wactlar.
\newblock 1995.
\newblock Informedia digital video library.
\newblock {\em Communications of the ACM}, 38(4):57--58.

\bibitem[\protect\citename{{Della~Pietra}, {Della~Pietra}, and
  Lafferty}1997]{DellaPietra:96a}
{Della~Pietra}, S., V.~{Della~Pietra}, and J.~Lafferty.
\newblock 1997.
\newblock Inducing features of random fields.
\newblock {\em IEEE Trans. on Pattern Analysis and Machine Intelligence},
  19(4):380--393, April.

\bibitem[\protect\citename{Hearst}1994]{Hearst:94}
Hearst, M.A.
\newblock 1994.
\newblock Multi-paragraph segmentation of expository text.
\newblock In Proceedings of the 32nd Annual Meeting of the ACL,
Las Cruces, NM.

\bibitem[\protect\citename{Jelinek \bgroup et al.\egroup }1991]{Jelinek:91a}
Jelinek, F., B.~Merialdo, S.~Roukos, and M.~Strauss.
\newblock 1991.
\newblock A dynamic language model for speech recognition.
\newblock In {\em Proceedings of the DARPA Speech and Natural Language
  Workshop}, pp. 293--295, February.

\bibitem[\protect\citename{Katz}1987]{Katz:87a}
Katz, S.
\newblock 1987.
\newblock Estimation of probabilities from sparse data for the langauge model
  component of a speech recognizer.
\newblock {\em IEEE Transactions on Acoustics, Speech and Signal Processing},
  ASSP-35(3):400--401, March.

\bibitem[\protect\citename{Kozima}1993]{Kozima:93}
Kozima, H.
\newblock 1993.
\newblock Text segmentation based on similarity between words.
\newblock in Proceedings of the 31st Annual Meeting of the ACL, 
Columbus, OH, pp. 286--288.

\bibitem[\protect\citename{Kozima and Furugori}1994]{Kozima:94}
Kozima, H. and T.~Furugori.
\newblock 1994.
\newblock Segmenting narrative text into coherent scenes.
\newblock {\em Literary and Linguistic Computing}, 9:13--19.

\bibitem[\protect\citename{Kuhn and {de Mori}}1990]{Kuhn:90}
Kuhn, R. and R.~{de Mori}.
\newblock 1990.
\newblock A cache-based natural language model for speech recognition.
\newblock {\em IEEE Trans. on Pattern Analysis and Machine Intelligence},
  12:570--583.

\bibitem[\protect\citename{Lau, Rosenfeld, and Roukos}1993]{Lau:93}
Lau, R., R.~Rosenfeld, and S.~Roukos.
\newblock 1993.
\newblock Adaptive language modeling using the maximum entropy principle.
\newblock In {\em Proceedings of the ARPA Human Language Technology Workshop},
  pages 108--113. Morgan Kaufman Publishers.

\bibitem[\protect\citename{Litman and Passonneau}1995]{Litman:95}
Litman, D.~J. and R.~J. Passonneau.
\newblock 1995.
\newblock Combining multiple knowledge sources for discourse segmentation.
\newblock In Proceedings of the 33rd Annual Meeting of the ACL,
Cambridge, MA.

\bibitem[\protect\citename{Neal}1992]{Neal:92}
Neal, R.
\newblock 1992.
\newblock Connectionist learning of belief networks.
\newblock {\em Artificial Intelligence}, 56:71--113.

\bibitem[\protect\citename{Reynar}1994]{Reynar:94}
Reynar, J.~C.
\newblock 1994.
\newblock In Proceedings of the 32nd Annual Meeting of the ACL,
student session, Las Cruces, NM.

\bibitem[\protect\citename{Youmans}1991]{Youmans:91}
Youmans, G.
\newblock 1991.
\newblock A new tool for discourse analysis: The vocabulary-management profile.
\newblock {\em Language}, 67:763--789.

\end{thebibliography}

\end{document}